%% file: ms.tex
\documentclass{article} % For LaTeX2e
\usepackage[final]{neurips_2019}

\input{math_commands.tex}

\usepackage{graphicx}
\usepackage{fancyhdr}
\usepackage{threeparttable}
\usepackage{xr}    
\usepackage{multirow}
\usepackage{siunitx}
\usepackage{wrapfig,lipsum,booktabs}
\usepackage{hyperref}
\usepackage{url}
\usepackage[utf8]{inputenc}
\usepackage{xcolor}
\usepackage{amsmath,amssymb}
\usepackage{algorithm}% http://ctan.org/pkg/algorithm
\usepackage{algpseudocode}% http://ctan.org/pkg/algorithmicx

\title{ROMark: A Robust Watermarking System Using Adversarial Training}
\author{%
  Bingyang Wen\\
  Department of ECE\\
  Stevens Institute of Technology\\
  Hoboken, NJ 07030 \\
  \texttt{bwen4@stevens.edu} \\
   \And
   Sergul Aydore \\
   Department of ECE\\
   Stevens Institute of Technology\\
   Hoboken, NJ 07030 \\
   \texttt{saydore@stevens.edu} \\}

\begin{document}

\maketitle

\begin{abstract}
The availability and easy access to digital communication increase the risk of copyrighted material piracy. In order to detect illegal use or distribution of data, digital watermarking has been proposed as a suitable tool. It protects the copyright of digital content by embedding imperceptible information into the data in the presence of an adversary. The goal of the adversary is to remove the copyrighted content of the data. Therefore, an efficient watermarking framework must be robust to multiple image-processing operations known as attacks that can alter embedded copyright information. Another line of research \textit{adversarial machine learning} also tackles with similar problems to guarantee robustness to imperceptible perturbations of the input. In this work, we propose to apply robust optimization from adversarial machine learning to improve the robustness of a CNN-based watermarking framework. Our experimental results on the COCO dataset show that the robustness of a watermarking framework can be improved by utilizing robust optimization in training.
\end{abstract}

\section{Introduction}
Digital watermarking as a tool for preventing copyright violation of data has been an active research field for decades \citep{cox2007digital}. Typically, a pattern of bits is embedded into a host image with no visible degradation to the original image. An ideal watermarking system should guarantee that the embedded watermarks are imperceptible and unremovable by malicious attacks. Therefore, robust watermarking systems in the presence of adversary have been developed to declare rightful ownership. In watermarking, image processing operations such as image enhancement, cropping, resizing, or compression can be regarded as attacks. As a result, the performance of watermarking systems is commonly measured by their robustness to these attacks.

Similar to the watermarking, the field of \textit{adversarial machine learning} also seeks to improve the robustness of neural networks in an adversarial environment \citep{kurakin2016adversarial}. Adversarial examples can be defined as specifically crafted inputs by an attacker to cause the neural network models to misbehave. This phenomenon was first observed by \cite{szegedy2013intriguing}. To mitigate this problem, the notion of adversarial training has been proposed. The basic idea is injecting adversarial examples into the training set at every step of training neural networks \citep{goodfellow2014explaining}.

Adversarial learning aims to minimize the adversarial risk as opposed to the traditional risk. Adversarial risk is the expected worst-case loss of each sample in some region around the sample point instead of the loss on each sample point. Hence, adversarial training can be formulated as the min-max or robust optimization problem where the task of inner maximization perturbs inputs within a region so that the loss is maximized and the outer minimization optimizes the parameters of the neural network so that the worst-case loss is minimized. This provides a more accurate estimate of the performance of the neural network operating in an adversarial environment.

Most recently, \cite{quiring2018forgotten} attempted to bring digital watermarking and adversarial machine learning together due to the similarities in defense and attack strategies in both fields. They provided a unified notation for black-box attacks in both fields to enable transferring concepts. In this work, we formulate a robust watermarking framework ROMark by employing the concepts from the robust optimization. Watermarking schemes typically contain two components: an \textit{encoder} and a \textit{decoder}. The encoder takes an image as well as a watermark message and produces a watermarked image. The decoder recovers the watermark from the watermarked image. Assuming both encoder and decoder are neural networks, an adversarial attack can be simulated at the output of the encoder. In our work, we apply a set of attacks at the output of the encoder and feed the worst-case attacked image to the decoder. We then optimize the parameters of both encoder and decoder.

\textbf{Related Work:} Using deep networks in watermarking frameworks has become popular most recently \citep{zhu2018hidden, mun2017robust, ahmadi2018redmark}. Among these, CNN-based HiDDeN \citep{zhu2018hidden} is the most relevant work to ours as it also uses adversarial training. HiDDeN achives robustness in two ways:  (i) by inserting a noise layer between the output of the encoder and the input of the decoder, and (ii) by adding an adversarial loss to the objective loss. However, HiDDeN does not solve the min-max optimization directly. 

\textbf{Our Contributions:} We adopt the architecture of the HiDDeN but compute the worst-case attacked image in the noise layer whereas HiDDeN's noise layer outputs attacked (adversarial) image by using a fixed set of parameters for the attacks. Our experiments on the COCO dataset demonstrate that our min-max formulation in training watermarking framework improves robustness to different types of image transformations.

%\section{Background}
%% Sergul: let's comment out these for now, may add back for a main conference submission
%\textbf{Robust Optimization:}
%
%Robust Optimization is an area of optimization theory which aims to obtain solutions which are stable under some level of uncertainty the data. The uncertainty has a deterministic and worst-case nature. The assumption is that the perturbations of the data can be drawn from specific sets $U_i$ called uncertainty sets. The uncertainty sets are often defined in terms of the type of the uncertainty and a parameter controlling the size of the uncertainty set. The Cartesian product of the sets $U_i$ is usually denoted by $U$.
%The goal in Robust Optimization is to obtain solutions which are feasible and well-behaved under any realization of the uncertainty from $U$; among feasible solutions, an optimal one would be such that has the minimal cost given the worst-case realization from $U$. Robust Optimization problems thus usually have a min-max formulation, in which the objective function is being minimized with respect to a worst-case realization of a perturbation.
%Inspire by robust optimization, we strengthen our watermarking system to malicious attacks in a way of training on the worst-case attacked images.
%
%\textbf{HiDDeN:}

\vspace{-0.3cm}
\section{Proposed approach: ROMark}
\vspace{-0.3cm}
Robust optimization aims to obtain solutions against the worst-case realizations of the data from a known uncertainty set. In the case of designing a robust watermarking system in an adversarial environment, the robust optimization formulation can be defined as solving two sub-problems: (i) obtaining the worst-case watermarked images that induce the largest decoding error within limits; (ii) optimizing parameters of the watermarking model on the worst-case watermarked images so that the loss of worst-case is minimized.

More formally, let $E_{\theta}$ parameterized by $\theta$ and $D_{\phi}$ parameterized by $\phi$ denote the encoder and the decoder of the watermarking framework, respectively. The encoder outputs the watermarked image $x^{wm}$ by embedding a binary secret message $m$ into a cover image $x$ so that $x^{wm} =  E_{\theta}(x, m)$. The watermarked images should perceptually look similar to the cover images. Therefore, the similarity distance between these can be characterized by the loss function $L_E(x, x^{wm})$ which typically measures the $\ell_2$ distance.
The decoder reconstructs message $\hat{m}$ that has the same shape as $m$ contained in the watermarked image $x^{wm}$: $\hat{m} = D_{\phi}(x^{wm})$. The similarity between $m$ and $\hat{m}$ indicates the success of the decoding process. We define a loss function $L_D$ to measure the difference between the embedded message and the reconstructed message from the decoder.
Hence the empirical objective function of our robust watermarking framework can be formulated as the min-max problem as follows:
\begin{equation}
\min_{\theta, \phi} \frac{1}{n} \sum^{n}_{i=1} \max_{x_i^{att} \in U_i}L_{D}\left(m_i, D_{\phi}\left(x_{i}^{att}\right)\right) + L_{E}\left(E_{\theta}\left(x_i, m_i\right), x_i \right)
\label{eqn:general_formular}
\end{equation}
where $U_i$ is the uncertainty set corresponding to the $i$-th image and $x_i^{att}$ is the corresponding simulated attacked image (or the adversarial example). 
\vspace{-0.3cm}
\subsection{Inner maximization: Obtaining Worst-case Attacked Images}
\vspace{-0.3cm}
Solving outer minimization in equation \ref{eqn:general_formular} requires access to the worst-case attacked images $x_i^{att}$. This maximization problem can be solved by finding the attacked image $x_i^{att}$ within a constraint set around $x_i^{wm}$ which maximizes the probability that the decoder fails to recover the watermarks. In digital watermarking, adversarial attacks are typically caused by image distortions such as \textit{crop}, \textit{image compression} and \textit{blurring}. Therefore, we define the images distorted by these attacks with varying severity levels as our uncertainty set. Let's assume there are $K$ image distortion functions $N_i$ where $i \in \{1, \cdots, K \}$ with corresponding severity level sets $S_i$. The worst-case attacked image can be defined as:
%%%%%%%%%%%%%%%%%%%%%%%%%%%%%%%%%%%%5
%One significant consideration for a digital watermarking model is the robustness to adversarial attacks caused by image distortions. Image distortions are some image processing operations done to the digital images while distributing or transmitting them, such as \textbf{Crop}, \textbf{Image Compression} and \textbf{ Blurry}, etc. These image processing operations can potentially cause a decoder to fail to recover embedded watermarks from the processed images. As a result, we can formulate these processed images (also known as attacked images) as an uncertainty set and obtain a robust solution for our watermarking model by robust optimization. There is usually a severity level that used to control the degree of processing the images, e.g., a quality parameter $Q \in (0,100)$ for JPEG compression to control the compression loss. The severity levels of these image distortion can be represented by a finite set of only a few elements. Consequently, we formulate the uncertainty set to be a set of attacked images by different types of image distortion with different severity levels. Saying we have several image distortion functions $N_1, N_2, ..., N_{i}$ that each takes a watermarked image $x_{wm}$ and a corresponding control parameter $s$ as input and outputs an attacked image $x_{att}$. $s$ is from a set of the severity levels of $S$ and $N_i$ are contained in a set $K$. Then the worst-case attacked images is obtained by:
\begin{equation}
x^{att*} = N^*(x^{wm}, s^{*})
\label{eqn:worst_case_attack}
\end{equation}
where $s^{*}$ and $N^{*}$ are obtained by:
\begin{equation}
s^{*}, N^{*} = \argmax_{N \in \{N_1, \cdots, N_K \},  s \in \{S_1, \cdots, S_K \}} L_D  \left(m, D_{\phi} \left(N(x^{wm}, s \right) \right) 
\label{eqn:worst_case_para}
\end{equation}
%\begin{equation}
%s^{*} = \argmax_{s \in S}\sum^{n}_{i=1}L(m_{i}, D_{\phi}(A(x_{wm}^{i}, s))) 
%\label{eqn:worst_case_para}
%\end{equation}
\vspace{-0.3cm}
\subsection{Outer minimization: Optimizing the Model Parameters}
\vspace{-0.3cm}
\label{sec:outer_min}
The goal of the outer minimization problem is to optimize the model parameters that minimizes the worst-case decoding loss. Reducing the worst-case loss offers a robustness guarantee that none of the considered attacks would induce a loss of large magnitute, i.e., successfully removes watermarks. %For watermarking models, both encoder and decoder should make effort for watermarking robustness by enhancing the embedding robustness of encoder and decoding ability of decoder. Consequently, we optimize both encoder and decoder to reduce the worst-case decoding loss in train-time.
Generally, after obtaining the worst-case attacked images, the outer minimization problem can be then represented as:
%\begin{equation}
%\min_{\theta, \phi} \frac{1}{n} \sum^{n}_{i=1} L_{D}\left(m_i, D_{\phi}\left(x_{i}^{att*}\right)\right) + L_{E}\left(E_{\theta}\left(x_i, m_i\right), x_i \right)
%\label{eqn:general_formular}
%\end{equation}
\begin{equation}
\min_{\theta, \phi} \frac{1}{n} \sum^{n}_{i=1} L_{D}\left(m_i, D_{\phi}\left(N^*(E_{\theta}\left(x_i, m_i\right), s^{*})\right)\right) + L_{E}\left(E_{\theta}\left(x_i, m_i\right), x_i \right)
\label{eqn:outer_general}
\end{equation}
Note that, $N^{*}$ in equation \ref{eqn:outer_general} should be differentiable to enable gradient derived from $L_{D}$ to backpropogate to encoder $E_{\theta}$. 

\begin{algorithm}[!b]
  \caption{Adversarial training of ROMark Combined}
  \begin{algorithmic}[l]
   \Require Batch size: $b$, Learning Rate: $\gamma_{\beta}$,$\gamma_{\theta}$,$\gamma_{\phi}$, Attack functions: $N_1,...,N_K$
   	\State{Randomly initialize the networks: $D_{\phi}$, $E_{\theta}$ and $C_{\beta}$.} 
   	\State{Randomly sample message batch $M$ of batch size $b$.}  
   	\State{Select K integers: $k_{1},...,k_{i},...,k_{K}$, where $K$ is the number of types of attacks and $\sum_{i=1}^{K}k_{i}=b$}
      \Repeat 
      	\State{Read minibatch $B=\{x_1,...,x_b\}$ from training set.}
      	\State{Generate the watermarked minibatch $B_{wm}=\{E_{\theta}(x_i,m_i): x_i \in B, m_i \in M\}$ }     	     \State{Separate the minibatch $B_{wm}$ into $K$ subsets $\{B^{1}_{wm},...,B^{K}_{wm}\}$ where each contains $k_{i}$ images}
      	\State{Load severity ranges of attacks: $S_1,...,S_K$} 
      	\For{i = 1, 2,..., K}
      	\State Search the worst-case $s^{*}_{i}$  by: $s^{*}_{i} = \argmax_{s \in S_i}\sum_{ x^{wm} \in B^{i}} L(m, D_{\phi}(N_i(x^{wm}, s))) $
        	\State Calculate the worst-case attacked image batch $B^{i}_{att} = \{N_{i}(x^{k}_{wm}, s^{*}_{i}): x^{k}_{wm} \in B^{i}_{wm} \}$ 
     	 \EndFor
      	\State{Generate attacked minibatch $B_{att}=\{B^{1}_{att} ,...,B^{K}_{att} \} $}
      	\State Feed $B_{att}$ into decoder, and then do one step training step:
      	\State \hspace{0.3cm}Updating discriminator C: \\
      	\hspace{1cm} $\beta_{t+1} =  \beta_{t} - \gamma_{\beta}  \sum_{x_i \in B, x^{wm}_{i} \in B_{wm}}\nabla_{\beta} A(x_{i},x^{wm}_{i})$ 
         \State \hspace{0.3cm}Updating $D_{\phi}$ and $E_{\theta}$: \\
      	\hspace{1cm} $\theta_{t+1} =  \theta_{t} - \gamma_{\theta}  \nabla_{\theta} J(B,M)$ and $\phi_{t+1} =  \phi_{t} - \gamma_{\phi}  \nabla_{\phi} J(B,M)$ 
     \Until{Training losses converged}
  \end{algorithmic}
       \label{alg:train}
\end{algorithm}

\vspace{-0.3cm}
\subsection{Overall Training}
\vspace{-0.3cm}
\label{sec:overall_train}
In this section, we present the details of overall training.  We use the mean squared error (MSE) for the loss at the decoder: 
%$L_{D}(m_i, x^{att*}_i) = \|m_i - D_{\phi}(x^{att*}_i)\|^2$. 
$L_{D}(m_i, \hat{m_i}) = \|m_i - \hat{m_i}\|^2$.
The loss for the encoder $L_E$ comprises the MSE loss between the watermarked and the cover image: $L_{EI}(x_i, x_i^{wm}) = \|x_i- x_i^{wm}\|^2$ and an adversarial loss for the watermarked image: $L_{EA}(x^{wm}_i) = \log(1-C_{\beta}(x^{wm}_i))$. $C_{\beta}$ is a discriminator network that is parameterized by $\beta$, which is trained by minimizing the loss $A(x_i, x^{wm}_i) = \log(1-C_{\beta}(x_i))+ \log(C_{\beta}(x^{wm}_i)) $. Hence, for a set of training samples $X$ and the secret messages $M$, the outer minimization problem can be re-written as:
%\min_{\theta, \phi}\sum^{m}_{i=1} \max_{s \in S_i}L(m^i, D_{\phi}(N(E_{\theta}(x^i, m^i), s))) + G(E_{\theta}(x^i, m^i), x^i)
%\begin{equation}
%\min_{\theta, \phi} J_{\theta,\phi}(X, M) 
%\label{eqn:outer_min}
%\end{equation}

\begin{equation}
\min_{\theta, \phi} \left[ J_{\theta,\phi}(X, M) = \sum_{i=0}^n L_{D}(m_i, D_{\phi}(x^{att*}_i)) + \lambda_I L_{EI}(x_i, E_{\theta}(x_i, m_i)) + \lambda_A L_{EA}(E_{\theta}(x_i, m_i)) \right]
\label{eqn:outer_min}
\end{equation}
where $\lambda_A$ and $\lambda_I$ control the relative weights of losses and $x^{att*}_{i}$ is obtained by equation \ref{eqn:worst_case_attack}.  Our Algorithm for training ROMark using combination of all attacks is given in Algorithm \ref{alg:train}. Note that, due to computational issues we are only optimizing $s$ for each $N_i$ instead of optimizing both $s$ and $N_i$. We are investigating the latter as future work.
\begin{table}[t!]
 \begin{center}
  \scalebox{0.60}{
    \begin{tabular}{ c  c  c  c c  c  c c  c}\toprule
      Model & Attack Type & Identity (no attack) &Crop & Cropout & Dropout & Gaussian Blur& JPEG Compression& Combined\\ \midrule
      \multirow{2}{*}{ROMark}& Range &-& $(0.1, 0.8)$&$(0.3, 0.9)$ &$(0.3, 0.9)$&$(1, 5)$ &$(50,100)$& \textit{Combination of all}\\\cmidrule{2-9}
      & Step Size &-& $0.1$&$0.1$ &$0.1$&$1$ &$10$&-\\ \midrule
      
      HiDDeN & Intensity &-& $0.3$&$0.3$ &$0.3$&$2$ & - & \textit{Combination of all} \\ \bottomrule
    \end{tabular}}
  \end{center}
 \caption{Parameter settings of noise layers used in training HiDDeN and ROMark models.} 
 \label{tab:noise setting}
 \end{table}
 
\vspace{-0.6cm}
\section{Implementation Details }
\vspace{-0.3cm}
We apply our ROMark and HiDDeN to the COCO dataset \citep{lin2014microsoft} ($10,000$ for training and $1000$ for testing) and evaluate the robustness to image processing attacks. We use \textit{peak signal-to-noise ratio (PSNR)} and the \textit{bit accuracy} to measure the performance. PSNR measures the amount of distortion in the encoded images so that high value of PSNR indicates better quality of the images. Bit accuracy is the ratio of correctly recovered bits to the total number of bits in the decoded watermarks. The embedded watermarks are randomly sampled binary vectors with length of 30. We use crop, cropout, dropout, Gaussian Blur and JPEG compression with various severity levels as attacks. Both ROMark and HiDDeN are trained with these specialized attacks and also combination of all. For a fair comparison, we use the same network architecture and hyperparameters for both ROMark and HiDDeN. The parameters of the attacks used in training are shown in Table \ref{tab:noise setting}.
%\paragraph{\textit{Metric}} 
%We use peak signal-to-noise ratio (PSNR) to measure the image distortion of the encoded images.
%We evaluate the decoding performance by using \textbf{bit accuracy}, which is the ratio of number of correctly recovered bits to the total number of bits of the decoded messages. We also use the worst-case bit accuracy (WCBA) when comparing between ROMark models that trained against combined attacks. Given a specific set of attacks and their corresponding ranges of severity levels (e.g., Table \ref{tab:noise setting}), WCBA is the lowest bit accuracy a ROMark can obtain from this set, which denotes the lower bound of robustness of a ROMark model. 
%
%\paragraph{\textit{Dataset}} 
%We randomly sample 10,000 training images and 1,000 testing images from COCO training set without overlap of each other. All of the images are center cropped to $128 \times 128$ and normalized to 0.5 mean and 0.5 standard deviation. 
%
%\paragraph{\textit{Implementation Details}} 
%We transform images into YUV channel for training and testing. The embedded watermarks are randomly sampled binary vectors with length of 30. The attacks set are constructed by different various attacks with different severity level. We refer readers to original paper of HiDDeN\citep{zhu2018hidden} to have more details about the architecture of watermarking system and adversarial attacks metioned in \ref{tab:noise setting}. The details of attacks set is shown below:

\vspace{-0.3cm}
\section{Experimental Results}
\begin{figure}[h!]
\centering
\includegraphics[scale=0.68]{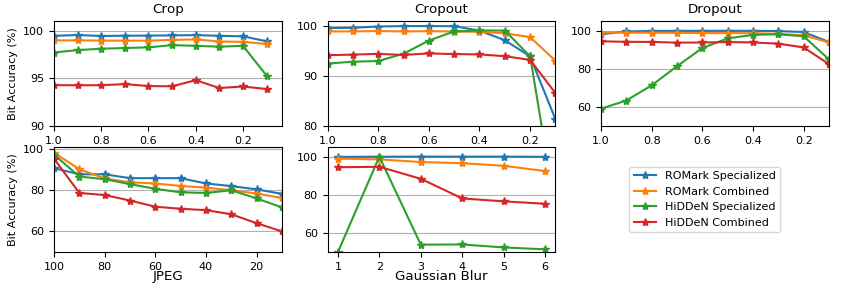}
\caption{Bit accuracy of ROMark models and HiDDeN models for various attacks and intensities. X-axis represents severity levels which increases from left to right.}
\label{fig:hiddenvsromark}
\end{figure}

\vspace{-0.3cm}
\label{sec:results}
%We experiment with both HiDDeN and ROMark that trained against specillized attacks and a combined attack. The attacks includes Crop, Cropout, Dropout, Gaussian Blur and JPEG compression. The combined attack comprises all of the attacks mentioned above. For more details about the effect of these attack, we refer reader to HiDDeN\citep{zhu2018hidden}. For a fair comparison, we train both ROMark model and HiDDeN model with the same network architecture and hyperparameters.
In Figure \ref{fig:hiddenvsromark}, we show the bit accuracy rates for both models under different attacks with different severity levels. When trained with the combination of all attacks, our ROMark Combined is more robust than HiDDeN Combined to all attacks at all severity levels. Using only the specialized attacks in training, our ROMark Specialized is more robust than the HiDDeN specialized for all attacks. Furthermore, HiDDeN Specialized yields higher accuracy under the attacks which were also used in training, i.e. overfits. Our ROMark, on the other hand, does not have the overfitting problem.

\vspace{-0.3cm}
%\textbf{Overfitting} We find that the specialized HiDDeN models can cause overfitting to the attack that applied in train-time. By examining the green lines in Figure \ref{fig:hiddenvsromark}, specialized HiDDeN models yield higher bit accuracy for the attack applied in training than that no attack implemented. By comparison, the specialized ROMark models (Blue Lines in Figure \ref{fig:hiddenvsromark}) do not have overfitting problems. Bit accuracies for specialized ROMark are stable around 99\% within the range of severity levels applied in training and gradually decrease as exceeding this range. This implies that our specialized ROMark models can yield more reliable and reasonable robustness: robustness is well guaranteed for seen attacks and gradually decreases as attacks become more severe.
\begin{table}[h!]
 \begin{center}
  \scalebox{0.70}{
    \begin{tabular}{ c c c c c c c}\toprule
      Model Type & Crop & Cropout & Dropout & Gaussian Blur& JPEG & Combined\\ \midrule
      HiDDeN & $24.32$ & $24.20$ & $24.20$ & $24.81$ & $23.57$ & $24.56$\\ \midrule
      ROMark & $26.78$ & $23.98$ & $26.58$ & $23.67$ & $27.70$ & $27.80$\\ \bottomrule
    \end{tabular}}
  \end{center}
 \caption{Average watermarking PSNR over 1000 testing images.} 
 \label{tab:imperceptibility}
 \end{table}
 
\vspace{-0.5cm}
\section{Conclusion}
We proposed a novel way to train a watermarking framework using the min-max formulation from robust optimization. The idea of minimizing the worst-case loss across several attacks makes the watermarking framework more robust to malicious attacks. Our experiments on the COCO dataset demonstrate that our min-max formulation in training watermarking framework improves robustness to different types of watermarking attacks. 

\clearpage 
\newpage
\bibliography{mybib}
\bibliographystyle{plainnat}

\end{document}

%% file: math_commands.tex
\usepackage{amsmath,amsfonts,bm}

\def\eqref#1{equation~\ref{#1}}
\def\1{\bm{1}}

\DeclareMathAlphabet{\mathsfit}{\encodingdefault}{\sfdefault}{m}{sl}
\SetMathAlphabet{\mathsfit}{bold}{\encodingdefault}{\sfdefault}{bx}{n}

\DeclareMathOperator*{\argmax}{arg\,max}